\documentclass[letterpaper,journal]{IEEEtran}
\usepackage{amsmath,amssymb,amsfonts}

\usepackage{placeins}
\usepackage{array}
\usepackage[caption=false,font=normalsize,labelfont=sf,textfont=sf]{subfig}
\usepackage{textcomp}
\usepackage{stfloats}
\usepackage{url}
\usepackage{verbatim}
\usepackage{graphicx}
\usepackage{tikz}
\usepackage{adjustbox}
\usepackage{cite}
\usepackage{xcolor}
\usepackage{tabu}
\usepackage{nomencl}

\usepackage{cancel}

\definecolor{darkergreen}{rgb}{0.0, 0.5, 0.0}
\definecolor{dgreen}{rgb}{0.3, 0.8, 0.0}

\usepackage{cancel}
\usepackage{stackengine}

\usepackage{multirow}
\definecolor{matlabgreen}{RGB}{119,172,48}
\usepackage{subcaption}
\usepackage{tikz}
\definecolor{darkblue}{rgb}{9,0,0}
\definecolor{lightyellow}{rgb}{1, 1, 0.8}
\definecolor{lightblue}{rgb}{0.9,0.9,1}
\usepackage{bm}
\usepackage{threeparttable}
\usepackage{caption}
\usepackage{tabularx}
\definecolor{lightgreen}{rgb}{0.75, 0.95, 0.75} 
\definecolor{lightgray}{rgb}{0.95, 0.95, 0.95}
\usepackage{soul}
\usepackage{booktabs}
\usepackage{listings}
\usepackage{xcolor} 

\definecolor{matlabblue}{RGB}{77,190,238}
\definecolor{matlabblue}{RGB}{77,190,238}
\definecolor{color1}{RGB}{113, 191, 234} 
\definecolor{color2}{RGB}{36, 123, 160} 
\definecolor{matlabpurple}{RGB}{126,47,142}
\definecolor{darkpink}{RGB}{255,19,166}

\lstdefinestyle{mystyle}{
    language=Matlab,
    basicstyle=\ttfamily\footnotesize,
    keywordstyle=\color{blue},
    commentstyle=\color{gray},
    stringstyle=\color{red},
    numbers=left,
    numberstyle=\tiny,
    numbersep=5pt,
    backgroundcolor=\color{white},
    frame=single,
    breaklines=true,
    showstringspaces=false,
    captionpos=b
}
\makenomenclature

\renewcommand{\nompreamble}{The following abbreviations and symbols will be used in the forthcoming equations.}

\usepackage{etoolbox}
\renewcommand\nomgroup[1]{%
  \item[\bfseries
    \ifstrequal{#1}{R}{DNN policy parameters}{%
    \ifstrequal{#1}{S}{RAC policy parameters}{%
    \ifstrequal{#1}{E}{MPD Parameters}{%
    \ifstrequal{#1}{A}{Abbreviations}{%
    \ifstrequal{#1}{Z}{Synthesis of DNN and RAC parameters}{%
    Other Symbols}}}}}%
  ]%
}

\begin{document}

\title{NMPC-Augmented Visual Navigation and Safe Learning Control for Large-Scale Mobile Robots}

\author{Mehdi Heydari Shahna, Pauli Mustalahti, and Jouni Mattila
\thanks{Funding for this research was provided by the Business Finland partnership project ``Future All-Electric Rough Terrain Autonomous Mobile Manipulators'' (Grant No. 2334/31/2022).}
\thanks{All authors are with the Faculty of Engineering and Natural Sciences, Tampere University, 33720 Tampere, Finland. (corresponding e-mail: mehdi.heydarishahna@tuni.fi)}}

\markboth{}%
{Shell \MakeLowercase{\textit{et al.}}: A Sample Article Using IEEEtran.cls for IEEE Journals}


\maketitle

\begin{abstract}
A large-scale mobile robot (LSMR) is a high-order multibody system that often operates on loose, unconsolidated terrain, which reduces traction. This paper presents a comprehensive navigation and control framework for an LSMR that ensures stability and safety-defined performance, delivering robust operation on slip-prone terrain by jointly leveraging high-performance techniques. The proposed architecture comprises four main modules: (1) a visual pose-estimation module that fuses onboard sensors and stereo cameras to provide an accurate, low-latency robot pose, (2) a high-level nonlinear model predictive control that updates the wheel motion commands to correct robot drift from the robot reference pose on slip-prone terrain, (3) a low-level deep neural network control policy that approximates the complex behavior of the wheel-driven actuation mechanism in LSMRs, augmented with robust adaptive control to handle out-of-distribution disturbances, ensuring that the wheels accurately track the updated commands issued by high-level control module, and (4) a logarithmic safety module to monitor the entire robot stack and guarantees safe operation. The proposed low-level control framework guarantees uniform exponential stability of the actuation subsystem, while the safety module ensures the whole system-level safety during operation. Comparative experiments on a 6,000 kg LSMR actuated by two complex electro-hydrostatic drives, while synchronizing modules operating at different frequencies.
\end{abstract}


\section{Introduction}

\IEEEPARstart{O}{ver} the next two decades, heavy vehicles are being transformed to large-scale mobile robots (LSMR) by integrating automation, Internet of Things 
(IoT), robotics, drones, and artificial intelligence. Vision-based LSMRs can operate continuously in unstructured environments, reduce accident risk, and improve precision in drilling, excavation, and heavy-load handling, thereby boosting productivity \cite{chang2022lamp}. For high-performance robotics, model-based control is an effective approach when an accurate system model is available \cite{abraham2020model}. Model-based term indicates how extensively controller design incorporates knowledge of the system’s dynamics. As a result, this approach, combined with vision-based algorithms, is now a key priority in both academia and industry because of its importance for autonomous navigation \cite{rizk2023model}.

Most modern model-based controllers require an accurate plant model, usually given as a transfer function or a state-space form \cite{he2025mimo}. However, compared with lightweight robots, LSMRs typically use complex mechanical chains, such as motors, couplings, and gearboxes that deliver transitional power to the ground-contacting wheels to create motion \cite{shahna2025robust}. Each link in the actuator chain has its own nonlinear dynamic characteristics, making the input-to-output behavior complex \cite{du2024hierarchical}. As a result, designing high-performance control in LSMRs requires building high-parameter models, which are analytically complex and computationally intensive \cite{bruni2020state}. To avoid relying on a fully accurate model, recent work often designs controllers around simplified dynamics \cite{archut2025systematic}, intentionally omitting wheel-terrain interactions and other hard-to-characterize behaviors \cite{liao2018model}. These simplifications can degrade control performance.
To address this challenge, machine learning model architecture can accurately approximate complex system dynamics by learning directly from available data and improve the control performance \cite{ostafew2016learning}. Unfortunately, deploying these high-performance learning methods on LSMRs is difficult since the trained models are not interpretable and are problematic in high-stakes settings. Furthermore, even small deviations in LSMRs from the defined trajectory can cause system instability and costly damage \cite{11175897}.
Moreover, these methods are constrained by their training distribution, bringing poor performance when test data fall outside the training domain \cite{li2025probing}, while for LSMRs on slip-prone terrain \cite{basri2022hybrid}, it is infeasible to gather data that spans all operating conditions for generalization, which increases the risk of immobilization \cite{arvidson2017mars}. For instance, the LSMR system in \cite{shahna2025antissdsd} requires human supervision to compensate for slip at individual wheels, thereby reducing the level of autonomy.
Slip is affected by vehicle factors (mass, speed, tire pressure) and terrain properties (sand compaction, rock friction, surface roughness). Therefore, multi-thousand-kilogram off-road LGMRs currently require an additional module to mitigate localization and path-tracking errors caused by excessive wheel slip \cite{basri2022hybrid}. To address this challenge, an efficient high-level control is needed to detect slip in real time and intelligently modify required wheel motions at the actuation level to return the robot to its desired path, while considering safety-defined constraints. Nonlinear model predictive control (NMPC) can update a reference motion online by pairing with a compatible algorithm with multiple sensor readings \cite{micheli2023nmpc}. In practice, real-time deployment of NMPC for LSMRs is difficult, since NMPC demands fast optimization, robust state estimation, and deterministic timing on embedded hardware, which limits horizon length and model fidelity and makes closed-loop execution challenging \cite{diehl2007stabilizing}. The multiple-shooting method introduced in \cite{bock1984multiple} can enable NMPC formulations to achieve robust optimal control while remaining suitable for real-time execution \cite{diehl2006fast}. However, if the control loop is slower than the system dynamics, the controller cannot reject fast disturbances, which can lead to instability, while for LSMRs, where navigation and control subsystems must be tightly synchronized, such online implementations are especially difficult. Finally, guaranteeing closed-loop stability and performance for the entire system of navigation and control in LSMRs remains challenging and has not yet been demonstrated in the literature.

This paper proposes a comprehensive architecture for an LSMR that addresses the aforementioned challenges, as follows: {1)} To circumvent the need for a complex multi-stage dynamic model of the LSMR, this paper employs a supervised deep neural network (SDNN) at the actuation level, while robust adaptive laws compensate for out-of-distribution disturbances to achieve high tracking performance. {2)} Following an in-depth assessment, a high-performance visual Simultaneous Localization and Mapping (SLAM) system with a stereo camera configuration was implemented and tuned for the LSMR under study. {3)} To mitigate localization errors caused by excessive wheel slip on unconsolidated terrain, this paper employs the multiple-shooting method, enabling NMPC to update the reference wheel motions for actuation-level control in real-time, while considering crucial constraints. {4)} A logarithmic safety module is integrated into the framework, guaranteeing operational safety and enabling the use of black-box methods. Hence, the proposed SDNN-based low-level control framework guarantees uniform exponential stability of the actuation subsystem, while the safety module ensures the whole system-level safety during operation. {5)} Performance improvements were validated in real-world experiments while synchronizing components that operate at different frequencies. To our knowledge, this is the first report of an LSMR combining visual SLAM, both high-level and low-level control, with a log-barrier safety supervisor, guarantees of actuation-level stability, and whole system safety.

The remainder of the paper is organized as follows. Section II develops the SDNN policy with adaptive controllers within a complex wheel-driven actuation mechanism to adjust real high-torque wheel motions to the required ones. Section III presents the implementation of a visual SLAM method, for accurate pose estimation during navigation. Section IV develops a high-level NMPC, which fuses sensor data to compensate wheel slip and refines the reference wheel motion based on the current robot obtained from the SLAM module. Section V provides the stability analysis and establishes uniform exponential stability. Section VI presents experimental validation of the overall framework for a 6,000-kg LSMR.

\begin{figure}[h!]
\hspace*{-0.0cm} 
\centering
\scalebox{1}{\includegraphics[trim={0cm 0.0cm 0.0cm 0cm},clip,width=\columnwidth]{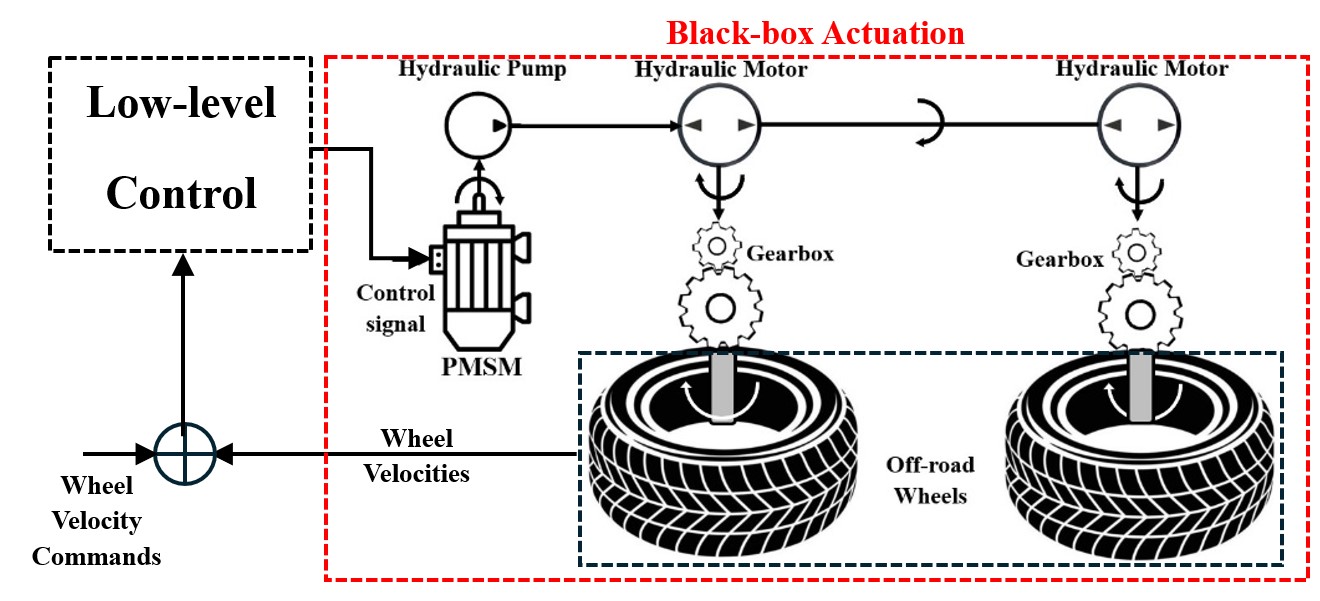}}
\caption{Electro-hydrostatic drive, as a black-box model.}
\label{dnn_contcomdsfsdf}
\end{figure}

\section{Low-Level Control: Robust Sdnn (Rsdnn)}
LSMRs employ complex actuation mechanisms to generate high forces, with multiple energy conversion stages (either electric or hydraulic, or both), gearing, and wheel-to-ground interaction, which makes system modeling challenging. However, to ensure the wheels track the commanded motion, high-performance control is required to adjust actuator signals in real time, and it depends on accurate modeling of this complex in-wheel actuation dynamics. For example, the studied LSMR, a 6000 kg skid-steering robot, uses a hybrid in-wheel architecture in which a permanent-magnet synchronous motor (PMSM) on each side drives a hydraulic pump that creates the pressure differential for two in-wheel hydraulic motors (one per wheel); torque is transmitted through a gear reduction to the wheels. In this paper, we consider the entire actuation mechanism as a black box, with only the input and output known and the internal behavior unknown; see Fig. \ref{dnn_contcomdsfsdf}.

\subsection{SDNN Policy to Model the Actuation Mechanism}
Since the two wheels on each side of the studied LSMR have the same velocity, we standardize the SDNN method's input to a per-side velocity command: $v_i=r \dot{\theta}_i$ for $i \in\{L, R\}$ where $\dot\theta_i \in \mathbb{R}$ is the angular velocity command of the $i$ side of the LSMR and $r \in \mathbb{R}$ is the radius of the wheels. SDNN's output is the PMSM RPM signal $u_{\mathrm{SDNN}_i} \in \mathbb{R}$. We can collect input/output data for the block-box mechanism as for each side, ramp the control input (actuator signal) $u_{\mathrm{SDNN}_i}$ from 0 to its rated limits in both directions within a safe operating range and under user supervision, while sensor readings log the corresponding wheel speeds. Then, we can use a feedforward SDNN, a multilayer perceptron (MLP) per side. The network has $L$ hidden layers with sizes $n_{\ell}(\ell=1, \ldots, L)$ where $n_0=1$ and $n_{L+1}=1$. For layer $\ell$,
$a^{(0)}=v_i, z^{(\ell)}=W^{(\ell)} a^{(\ell-1)}+b^{(\ell)}, a^{(\ell)}=\varphi^{(\ell)}(z^{(\ell)})$
where $W^{(\ell)} \in \mathbb{R}^{n_{\ell} \times n_{\ell-1}}$ is the weight matrix, $\varphi^{(\ell)}$ is the activation function (e.g. hyperbolic
tangent), $b^{(\ell)} \in \mathbb{R}^{n_{\ell}}$ is the corresponding bias vector, and pre-activation (affine transformation) is denoted as $z^{(\ell)}, a^{(\ell)} \in \mathbb{R}^{n_{\ell}}$. The output layer is
$
\varphi^{(L+1)}(z)=z$ where $u_{\mathrm{SDNN}_i}=a^{(L+1)}=z^{(L+1)}
$. With a mini-batch of $P$ samples, stack them into the $1 \times P$ input row $\mathcal{V}_{i}=[v_{i}^{(1)}| v_{i}^{(2)} | \cdots |v_{i}^{(P)}] \in \mathbb{R}^{1 \times P}$.
The layerwise forward pass is $A^{(\ell)}=\varphi^{(\ell)}(W^{(\ell)} A^{(\ell-1)}+b^{(\ell)} \mathbf{1}^T)$ where $A^{(0)}=\mathcal{V}_{i}$
where $\mathbf{1} \in \mathbb{R}^P$ is a column of ones.
Let $\psi=\{W^{(\ell)}, b^{(\ell)}\}_{\ell=1}^{L+1}$ collect all trainable parameters. 
We quantify the mismatch between predictions $u_{\mathrm{SDNN}_i}$ and targets $t_{p_i}$ using the error loss
$E_{\mathrm{MSE}}(\psi)=\frac{1}{P} \sum_{p=1}^P\|{u}_{i}^{(p)}-t_{p_i}^{(p)}\|_2^2 =\frac{1}{P}\|{U}_{i}-T_{p_i}\|_F^2$
where $U_{i}$ and $T_{p_i}$ are $n_{L+1} \times P$ matrices stacking the network outputs and targets, respectively, and $\|.\|_F$ denotes the Frobenius norm.
To reduce iterations versus plain gradient descent, we use Levenberg–Marquardt (LM) backpropagation: cast training as nonlinear least squares, approximate the Hessian via Gauss–Newton, and stabilize with a damping term \cite{zhou2018levenberg}. Let $w \in \mathbb{R}^{N_w}$ stack all learnable parameters: for each layer, concatenate $W^{(\ell)}$ with $b^{(\ell)}$. Its length is
$
N_w=\sum_{\ell=1}^{L+1}(n_{\ell} n_{\ell-1}+n_{\ell})
$.
Stack the per-sample errors into
$
\xi_{p_i}(w)=[
\xi_{p_i}^{(1)},
\xi_{p_i}^{(2)},
\ldots,
\xi_{p_i}^{(P)}]^T \in \mathbb{R}^{m P}$ where $m=n_{L+1}=1$.
The mean-squared error (scaled by $1 / P$) is
$
E_{\mathrm{MSE}}(w)=\frac{1}{P}\|\xi_{p_i}(w)\|_2^2=\frac{1}{P} \xi_{p_i}(w)^{\top} \xi_{p_i}(w)
$. Thus, minimizing $E_{\text {MSE }}(w)$ is a nonlinear least-squares problem.
Let us define the Jacobian as
$J_{k, j}(w)=\frac{\partial s_k(w)}{\partial w_j}$
where $s_k$ is the $k$-th entry of the stacked error vector $\xi_{p_i}$, and the exact gradient of $E_{\mathrm{MSE}}$ is
$\nabla E_{\mathrm{MSE}}(w)=\frac{2}{P} J(w)^T \xi_{p_i}(w)$.
Since one term of the Hessian function contains second derivatives, computing the exact Newton step $\Delta w= -H(w)^{-1} \nabla E(w)$ is costly. Gauss-Newton instead uses $H(w) \approx \frac{2}{P} J(w)^T J(w)$, giving
$\Delta w_{\mathrm{GN}}=-(J^T J)^{-1} J^T \xi_{p_i}$.
Adding Tikhonov damping to mitigate the near-singularity of $J^T J$, we have
$\Delta w=-(J^T J+\mu I)^{-1} J^T \xi_{p_i}$.
When $\mu$ is large, $\mu I$ dominates and $\Delta w \approx-\frac{1}{\mu} J^{\top} \xi=-\eta \nabla E_{\mathrm{MSE}}$ with $\eta=1 / \mu$; as $\mu \to 0$, it recovers the Gauss-Newton step. Thus, LM smoothly moves between stable-but-slow GD and fast-but-less-stable GN.
At iteration $b$ : compute $J_b, \xi_b$ at $w_b$; choose $\mu_b$; solve $(J_b^{\top} J_b+\mu_b I) \Delta w_b=-J_b^{\top} \xi_b$; set $w_{\text {cand }}=w_b+ \Delta w_b$ and evaluate $E_{\mathrm{MSE}}$. If improved, accept $w_{b+1}=w_{\text {cand }}$ and decrease $\mu(\mu_{b+1}=\mu_b / \beta, \beta \approx 10)$; otherwise, keep $w_{b+1}=w_b$ and increase $\mu(\mu_{b+1}=\beta \mu_b)$.

\subsection{Trained SDNN Model Augmented by Adaptive Control}
After acceptably passing evaluations, the LM-trained SDNN model is ready in real-time to accept velocity commands $v_i=r \dot{\theta}_i$ and outputs the predicted control signal $u_{\text{SDNN}_i}$, which serves as the actuation modeling term in the low-level control framework corresponding to the wheel reference velocity $\dot{\theta}_i$. It means that if, during operation, the robot’s conditions remain within the training data distribution, for each side of the robot, we ideally have the second-order dynamics, as
\begin{equation}
\begin{aligned}
\small
\label{3adasd5}
&A_i \ddot{\theta}_{i} (t) = u_{\text{SDNN}_i} (\dot{\theta}_{i}, t)  + F_i(\dot{\theta}_i, t)
\end{aligned}
\end{equation}
where it is a nominal condition and $\dot{\theta}_{\text{msr}_i} = \dot{\theta}_i$ and $F_i \in \mathbb{R}$ denotes the nonlinear, unknown mapping determined by the robot's actuation mechanism under nominal conditions as well as $A_i \in \mathbb{R}^+$. In LSMRs, however, out-of-distribution conditions and external disturbances may occur. Under such conditions, applying the SDNN-based policy yields measured wheel velocities that deviate from the reference, producing the non-zero error $e_i(t)=\dot{\theta}_{\mathrm{msr}_i}-\dot{\theta}_i$. Accordingly, we extend the ideal model in Eq. \eqref{3adasd5} to the real-world form
\begin{equation}
\begin{aligned}
\small
\label{3asdas5}
&A_i (\ddot{\theta}_{i} + \dot{e}_i) = u_i(t)  + F_i(\dot{\theta}_i,t) + d_i(\dot{\theta}_i, e_i,t)
\end{aligned}
\end{equation}
where $d_i(.) \in \mathbb{R}$ captures unmodeled dynamics and disturbances, and $u_i(t)$ is the control signal which is required to be designed. Thus from \eqref{3adasd5} and \eqref{3asdas5}, we finally have
\begin{equation}
\begin{aligned}
\small
\label{35}
\dot{e}_{i} = A^{-1}_i [ u_i(t) - u_{\text{SDNN}_i}(\dot\theta_i) + d_i(\dot{\theta}_i, e_i,t)]
\end{aligned}
\end{equation}

\textit{Assumption II.1.} Consistent with established results in the robust control literature \cite{tan2025fixed}, assume that the functional control gain $A_i$ and the dynamical functions $F_i(.)$ and $d_i (.)$ are locally Lipschitz, continuous, and bounded.

We now design the developed control policy $u_i (.)$ that is valid under all operating conditions of the actuation mechanism, both nominal and perturbed, as follows

\begin{equation}
\begin{aligned}
\small
\label{vbfvfvf39}
u_{i} (t) = u_{\text{SDNN}_i} (\dot\theta_i) -  \frac{1}{2} \epsilon_i e_i - \gamma_i e_i \log^2 (\frac{{O}}{O-{E}}) \hat{\chi}_i
\end{aligned}
\end{equation}
where $\gamma_i \in \mathbb{R}^+$ and $\epsilon_i \in \mathbb{R}^+$, and $\hat{\chi}_i$ is the proposed adaptive law which is defined, as
\begin{equation}
\begin{aligned}
\small
\label{3aadasdasd2}vbfvfvf39
\dot{\hat{\chi}}_i (t) = -\delta_i \hat{\chi}_i+\gamma_i e^2_i \log^2 (\frac{{O}}{O-{E}})
\end{aligned}    
\end{equation}
where $\delta_i \in \mathbb{R}^+$, and $\hat{\chi}_i(t_0) \in \mathbb{R}^+$. Following \cite{tee2009barrier, shahna2025robust}, and as shown in Eqs. \eqref{vbfvfvf39} and \eqref{3aadasdasd2}, we define within the proposed low-level control framework a logarithmic barrier function that serves as the safety module for the entire system. 
Here, $E(t)$ denotes the robot pose error, which is defined as
\begin{equation}
\begin{aligned}
\small
\label{3asdsad2}
{E} = \sqrt{||\bm{x}_\text{msr}(t) - \bm{x}^r(t)||^2}
\end{aligned}    
\end{equation}
where $\boldsymbol{x}^r(t) \in \mathbb{R}^3$ is the reference pose trajectory and $\boldsymbol{x}_{\mathrm{msr}}(t) \in \mathbb{R}^3$ is the measured robot pose. We impose the threshold safety constraint $0 \leq E(t)<O$, where $O>0$.
 
\textit{Remark II.1.} As shown in Eqs. (\ref{3adasd5}--\ref{3asdsad2}), if the initial robot pose satisfies $E(t_0)<O(t_0)$, then during operation the tracking error is enforced to satisfy $E(t)<O(t)$, for all $t$. According to this, logarithmic safety module, if $E(.)$ increases to the predefined safety bound $O(.)$, numerical singularities may occur and execution can halt with warnings (e.g., "Infinity or NaN value encountered") when computed values exceed numerical limits \cite{shahna2025robust}. The safety supervisor can run a clear state machine that caps speed near the barrier, executes a deterministic braking profile when limits are crossed, and latches to a Safe stop on E-stop. 

\textit{Assumption II.2.} Required trajectory $\boldsymbol{x}^r(t)$ is a differentiable and continuous on the manifold $S E(2)$.

\section{Vision-based Robot Pose}
As shown in Eq. \eqref{3asdsad2}, the designed safety module within the low-level control framework is required the robot pose estimation $\bm{x}_\text{msr}(t)$ in real time.
For off-road LSMRs, vision-based pose estimation provides high-rate localization and environmental perception, where Global Navigation Satellite System (GNSS)-based pose estimation is degraded by propagation impairments \cite{furgale2010visual}. In addition, visual SLAM is a preferable GNSS-free architecture since it enables re-localization and drift correction through loop closure, and its mapping can later be extended to obstacle avoidance and other functions supporting long-term navigation in unknown environments \cite{cadena2017past}. ORB-SLAM3 is a mature open-source system, presented in \cite{campos2021accurate}. It demonstrates real-time stereo visual SLAM in indoor and outdoor settings, provides robust loop closure, and achieves top accuracy on EuRoC \cite{burri2016euroc} and TUM-VI \cite{schubert2018tum}. Thus, this paper utilizes ORB-SLAM3 for the studied LSMRs as the pose estimation module since it is accurate and robust on standard benchmarks. It includes loop closure and map fusion via DBoW2, with relocalization using EPnP. ORB-SLAM3 is a tightly integrated, keyframe- and feature-driven visual SLAM system that estimates pose via maximum a posteriori (MAP) methods with multiple temporal data-association strategies. It also supports our stereo input with either pinhole or fisheye camera models. ORB-SLAM3 can be implemented in C++, using several open-source libraries. It takes camera frames, runs a tracking thread that extracts ORB features, estimates the current pose from the last frame or via relocalization or map initialization, maintains alignment to the local map, and decides when to create a new keyframe. An Atlas manages multiple maps, with one active map and additional non-active maps, each holding keyframes, mappoints, a covisibility graph, and a spanning tree for multi-session reuse and later merging. A DBoW2 keyframe database provides a visual vocabulary and recognition index for fast place recognition and relocalization. The local mapping thread inserts new keyframes, culls recent mappoints and redundant keyframes, creates new points, performs local bundle adjustment, and scale refinement. Loop closing and map merging use database queries to detect places, compute SE3 or Sim3 constraints, fuse loops, optimize the essential graph, and weld or merge separate maps. After loop correction or merging, a full bundle adjustment refines the entire map \cite{campos2021accurate}. Finally, the state $\mathcal{S}$ contains the camera pose as $\mathbf{T}= [\mathbf{R}, \boldsymbol{x}_\text{msr}] \in S E(3)$ where
$\boldsymbol{x}_\text{msr}\ \in \mathbb{R}^3$ is the pose vector and $\mathbf{R} \in S O(3)$ is the rotation matrix. Section VI will present the experimental implementation and parameter tuning of the ORB-SLAM3 module for the studied LSMR.

\section{High-Level Control}

As shown in Fig. \ref{dnn_contcomdsfsdf}, the actuation of the LSMR requires wheel-velocity commands for the robot to execute a specific motion task. However, these commands must be updated in real time, since the LSMR operates on slip-sensitive terrain. Using the measured robot pose $\boldsymbol{x}_\text{msr}$ (see Section V), we design a high-level controller that updates wheel commands in real time to compensate for slip-induced deviations from the reference pose trajectory, subject to motion constraints. As the high-level control, this paper uses the multi-shooting NMPC proposed in \cite{paz2025real}, pairing with a compatible algorithm with multiple sensor readings for genuine real-time performance.
In the method, a transcription phase [13] converts an infinite optimal control problem (OCP) into a nonlinear programming problem (NLP), executed each iteration to achieve real-time rates and synchronized sensing. Assume the robot body frame is located at the geometric center of the wheel frames. Expressed in the body frame, the robot's twist is
$
\boldsymbol{\nu}^{(b)}=[\begin{array}{lll}
v_x & v_y & \omega_z
\end{array}]^{\top} \in \mathbb{R}^3,
$
stacking linear velocities along $x$ and $y$ and the angular velocity about $z$. Define the vector $\dot{\boldsymbol{\theta}}=[\begin{array}{ll}\dot{\theta}_R & \dot{\theta}_L\end{array}]^{\top} \in \mathbb{R}^2$, including right- and left-side velocities of the robot. The local frames are derived from the first-order kinematic map \cite{kozlowski2004modeling}, as
$
\boldsymbol{\nu}^{(b)}=\boldsymbol{J} \dot{\boldsymbol{\theta}}
$
where the Jacobian matrix is defined, as

\begin{equation}
\begin{aligned}
\small
\label{1212}
\boldsymbol{J}=\frac{r}{2}\left[\begin{array}{ccc}
1 & 0 & 1 / c \\
1 & 0 & -1 / c
\end{array}\right]^{\top}
\end{aligned}    
\end{equation}
where $c$ is half the lateral track width and $r$ is the wheel radius.

\textit{Assumption IV.1.} Commonly in mobile robots, the lateral velocity is assumed zero, i.e., $v_y=0$, thus the body frame $b$ is constrained to move tangentially to the path.

Using screw theory, where the Lie group $SE(2)=\mathbb{R}^2\times SO(2)$ captures all planar rigid motions between frames, the motion state of the body frame $b$ relative to $W$ is $\boldsymbol{x}_W^b=\begin{bmatrix}\boldsymbol{p}_W^b\ \alpha_W^b\end{bmatrix}\in\mathbb{R}^3$. $\boldsymbol{p}_W^b \in \mathbb{R}^2$ and $\alpha$ are the position and orientation, and $S O(2)$ and $S E(2)$ are the Lie group for orthogonal and Euclidean matrices. $\boldsymbol{x}_W^b \in \mathbb{R}^{3 \times 3}$ can be obtained from the homogeneous transform $\boldsymbol{G}_W^b \in \mathbb{R}^{3 \times 3}$, as

\begin{equation}
\begin{aligned}
\small
\label{1212}
\boldsymbol{G}_W^b=\left[\begin{array}{ll}
\boldsymbol{R}_W^b & \boldsymbol{p}_W^b \\
\mathbf{0} & 1
\end{array}\right]
\end{aligned}    
\end{equation}

As $\boldsymbol{\nu}^{(b)}$ belongs to $S E(2)$, it can be carried to the inertial frame $W$ via the adjoint operator $\operatorname{Ad}_G: S E(2) \to \mathbb{R}^{3 \times 3}$ \cite{murray2017mathematical}. Thus, viewing $\boldsymbol{\nu}^{(W)}$ as the pose time derivative
$\dot{\boldsymbol{x}}$ yields the velocity-level kinematic model
\begin{equation}
\begin{aligned}
\small
\label{121sadasd2}
\dot{\boldsymbol{x}}=\operatorname{Ad}_{G_W^b} \boldsymbol{\nu}^{(b)}=\operatorname{Ad}_{G_W^b} \boldsymbol{J} \dot{\boldsymbol{\theta}} = \boldsymbol{f}(\boldsymbol{x}, \dot{\boldsymbol{\theta}})
\end{aligned}    
\end{equation}

We formulate a tracking OCP for the reference trajectory $\boldsymbol{x}^r(t)$ over $t\in[0,T]$ using NMPC with input regularization. The objective is
\begin{equation}
\label{eq:ocp-cost}
\min_{\dot{\boldsymbol{\theta}}(\cdot)} \frac{1}{2}\int_{0}^{T}\dot{\boldsymbol{\theta}}(t)^{\top}\dot{\boldsymbol{\theta}}(t),dt .
\end{equation}
This optimization is subject to the system model in \eqref{121sadasd2}, as
\begin{equation}
\label{eq:ocp-cost}
\begin{aligned}
\text{s.t.}\;&
\left\{
\begin{aligned}
\boldsymbol{x}(t) &= \boldsymbol{x}^r(t),\\
\dot{\boldsymbol{x}}(t) &= \dot{\boldsymbol{x}}^r(t),\\
\dot{\boldsymbol{x}}_{\text{min}} &\le \dot{\boldsymbol{x}}(t) \le \dot{\boldsymbol{x}}_{\text{max}},\\
\dot{\boldsymbol{\theta}}_{\text{min}} &\le \dot{\boldsymbol{\theta}}(t) \le \dot{\boldsymbol{\theta}}_{\text{max}}.
\end{aligned}\right
.
\end{aligned}
\end{equation}

where $\dot{\boldsymbol{x}}_{\text{min}}$ and $\dot{\boldsymbol{x}}_{\text{max}}$ set the lower and higher bounds on the robot velocities. Similarly, $\dot{\boldsymbol{\theta}}_{\text{min}}$ and $\dot{\boldsymbol{\theta}}_{\text{max}}$ set the angular wheel velocity bounds. To convert the OCP into an NLP for NMPC with horizon $N \in \mathbb{N}$, introduce a time grid $t_k$ for $k=0, \ldots, N$ with step $\Delta t=t_{k+1}-t_k$. Let $\boldsymbol{x}_k$ and $\dot{\boldsymbol{\theta}}_k$ denote the state and control at step $k$. The continuous model \eqref{121sadasd2} are sampled as $\dot{\boldsymbol{x}}_k=\boldsymbol{f}(\boldsymbol{x}_k, \dot{\boldsymbol{\theta}}_k)$ and advanced via $
\hat{\boldsymbol{x}}_{k+1}=\text {int}(\boldsymbol{x}_k, \dot{\boldsymbol{x}}_k, \Delta_t)
$ where $\mathrm{int}(\cdot)$ denotes a numerical integrator (e.g., Euler, Runge–Kutta) used to approximate the next state $\hat{\boldsymbol{x}}_{k+1}$.
Stacking all states and controls over the horizon yields the decision vector $\boldsymbol{z}\in\mathbb{R}^{5N+3}$, as

\begin{equation}
\label{eq:ocp-cost}
\small
\begin{aligned}
\begin{aligned}
\boldsymbol{z}=[\begin{array}{lllllll}
\boldsymbol{x}_0^{\top} \hspace{0.2cm} \dot{\boldsymbol{\theta}}_0^{\top} \cdots \hspace{0.2cm} \boldsymbol{x}_k^{\top} \hspace{0.2cm} \dot{\boldsymbol{\theta}}_k^{\top}  \cdots  \boldsymbol{x}_N^{\top}
\end{array}]^{\top} 
\end{aligned}
\end{aligned}
\end{equation}
Accordingly, at the grid instants $t_k$ and $t_N$, define a stage cost $L_k(\cdot)$ and a terminal cost $L_N(\cdot)$, each aggregating multiple objectives, as
\begin{equation}
\label{eq:ocp-cost}
\small
\begin{aligned}
L_k(\boldsymbol{x}_k,\boldsymbol{x}_k^r,\dot{\boldsymbol{\theta}}_k)
&= \|\boldsymbol{x}_k-\boldsymbol{x}_k^r\|_{Q_x}^2
 + \|\dot{\boldsymbol{x}}_k-\dot{\boldsymbol{x}}_k^r\|_{Q_{\dot{x}}}^2
 + \|\dot{\boldsymbol{\theta}}_k\|_{\bar{R}}^2, \\
L_N(\boldsymbol{x}_N,\boldsymbol{x}_N^r)
&= \|\boldsymbol{x}_N-\boldsymbol{x}_N^r\|_{Q_{xN}}^2
 + \|\dot{\boldsymbol{x}}_N-\dot{\boldsymbol{x}}_N^r\|_{Q_{\dot{x}N}}^2
\end{aligned}
\end{equation}
At each grid time $t_k$, the stage cost is quadratic: $\bar{R} \in \mathbb{R}^{2 \times 2}$ weights control effort, while $Q_x, Q_{\dot{x}} \in \mathbb{R}^{3 \times 3}$ weight the tracking errors w.r.t. $\boldsymbol{x}_k^r$ and $\dot{\boldsymbol{x}}_k^r$. The terminal matrices $Q_{x N}, Q_{\dot{x} N} \in \mathbb{R}^{3 \times 3}$ impose the same penalties at $t_N$.
A state-feedback law is posed as an optimization problem initialized at $(t_0)$ using the measured state $\boldsymbol{x}_{\mathrm{msr}}$ and measured wheel velocities $\dot{\boldsymbol{\theta}}_{\mathrm{msr}}$, as
\begin{equation}
\label{eq:ocp-cost}
\small
\begin{aligned}
\underset{\boldsymbol{z}}{\arg \min } \hspace{0.2cm} J=\frac{1}{2} \sum_{k=1}^{N-1} L_k(\boldsymbol{x}_k, \boldsymbol{x}_k^r, \dot{\boldsymbol{\theta}}_k)+L_N(\boldsymbol{x}_N, \boldsymbol{x}_N^r)
\end{aligned}
\end{equation}
where
\begin{equation}
\label{eqocpcasdsaost}
\small
\begin{aligned}
\text{s.t.}\;&
\left\{
\begin{aligned}
&\boldsymbol{x}_0  =\boldsymbol{x}_{\mathrm{msr}}, \hspace{0.1cm} \dot{\boldsymbol{\theta}}_0=\dot{\boldsymbol{\theta}}_{\mathrm{msr}} \\
&\boldsymbol{x}_{k+1}  =\hat{\boldsymbol{x}}_{k+1} \\
&\boldsymbol{x}_{\min }  \leq \boldsymbol{x}_k \leq \boldsymbol{x}_{\max } && k=0 \cdots N \\
&\dot{\boldsymbol{\theta}}_{\min } \leq \dot{\boldsymbol{\theta}}_k \leq \dot{\boldsymbol{\theta}}_{\max } && k=0 \cdots N-1 \\
& \ddot{\boldsymbol{\theta}}_{\min } \leq \ddot{\boldsymbol{\theta}}_k \leq  \ddot{\boldsymbol{\theta}}_{\max } &&k=0 \cdots N-1
\end{aligned}\right
.
\end{aligned}
\end{equation}

\begin{figure}[h!]
\hspace*{-0.0cm} 
\centering
\scalebox{1}{\includegraphics[trim={0cm 0.0cm 0.0cm 0cm},clip,width=\columnwidth]{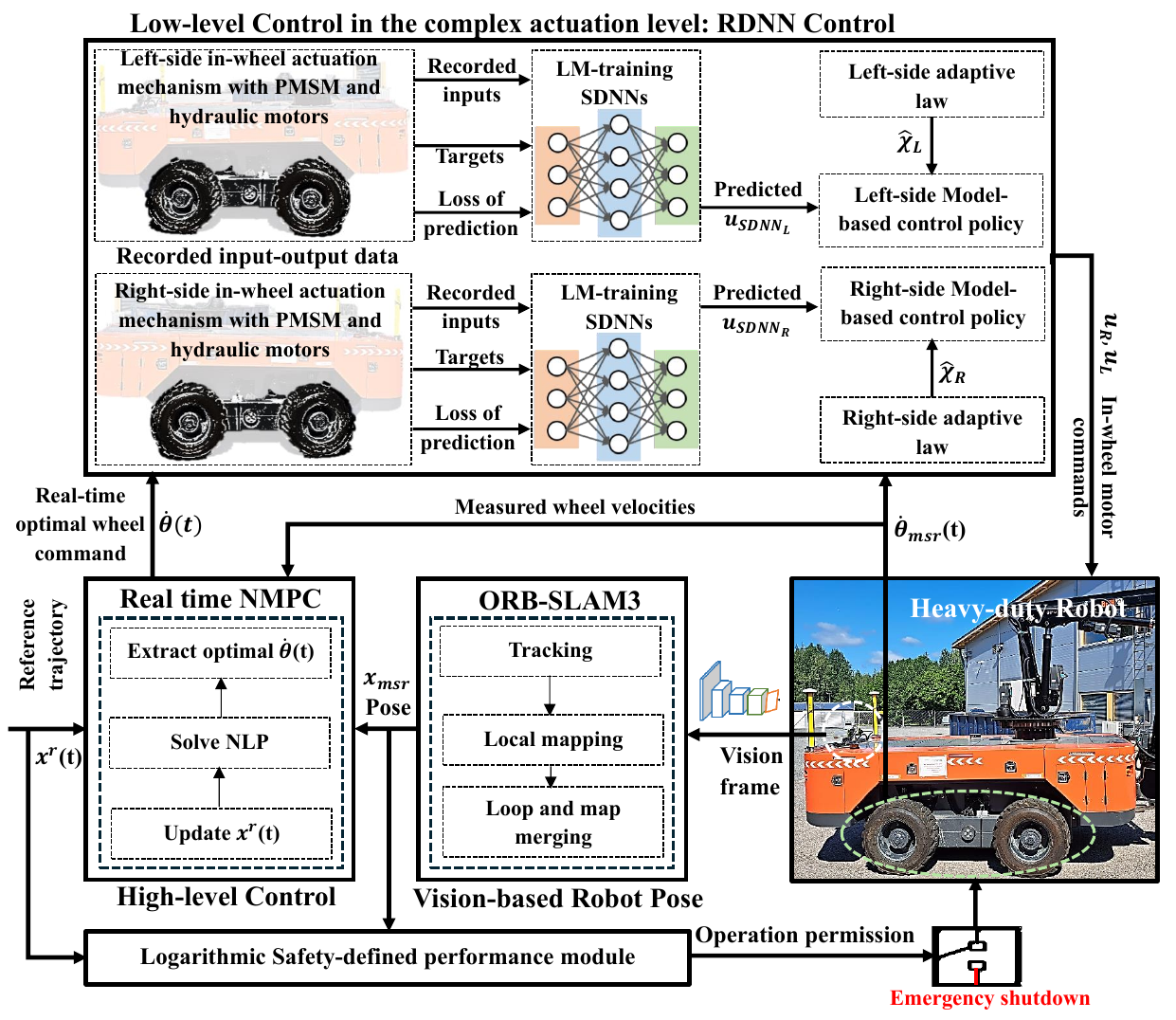}}
\caption{The proposed control framework for the studied LSMR.}
\label{i9sasddadado_synthesnbfgfhfhnfh}
\end{figure}

$J$ aggregates stage and terminal penalties. First constraint on Eq. \eqref{eqocpcasdsaost} fixes the initial state and input to the measured values, closing the loop. The second one enforces multiple-shooting consistency by matching decision states to the integrated dynamics. The rest ones impose box bounds on states $\boldsymbol{x}_k$, inputs $\dot{\boldsymbol{\theta}}_k$, and input rates $\ddot{\boldsymbol{\theta}}_k$ via given minima/maxima. The whole proposed control framework including different module for the studied LSMR is shown in Fig. \ref{i9sasddadado_synthesnbfgfhfhnfh}.

\section{Stability Analysis}
If the robot is on the running mode, based on the logarithmic safety module shown in Fig. \ref{i9sasddadado_synthesnbfgfhfhnfh} and Eqs. \eqref{vbfvfvf39} and \eqref{3aadasdasd2}, the robot is within the safety-defined constraint $E(t) < O$. Following this, define a Lyapunov function, as
\begin{equation}
\begin{aligned}
\small
\label{35}
\bar{V}=\frac{A_R}{2} e^2_R+ \frac{A_L}{2} e^2_L +  \frac{1}{2} \hat{\chi}_R^2 + \frac{1}{2} \hat{\chi}_L^2
\end{aligned}
\end{equation}
By differentiating \eqref{35}, and inserting $e_i$, we obtain
\begin{equation}
\begin{aligned}
\small
\label{37}
\dot{\bar{V}}= & e_R [u_R - u_{\text{SDNN}_R} + d_R]+ \hat{\chi}_R \dot{\hat{\chi}}_R \\
&+ e_L [u_L - u_{\text{SDNN}_L} + d_L(t)]+ \hat{\chi}_L \dot{\hat{\chi}}_L
\end{aligned}
\end{equation}
After inserting \eqref{vbfvfvf39}, we obtain
\begin{equation}
\begin{aligned}
\small
\label{37}
\dot{\bar{V}}= & - \frac{1}{2} \epsilon_R e^2_R - \gamma_R e^2_R [\log (\frac{{O}}{O-{E}})]^2 \hat{\chi}_R + e_R d_R\\
&+ e_R ({{u_{\text{SDNN}_R}-u_{\text{SDNN}_R}}})+ \hat{\chi}_R \dot{\hat{\chi}}_R\\
&- \frac{1}{2} \epsilon_L e^2_L - \gamma_L e^2_L [\log (\frac{{O}}{O-{E}})]^2 \hat{\chi}_L + e_L d_L\\
&+ e_L ({{u_{\text{SDNN}_L}-u_{\text{SDNN}_L}}})+ \hat{\chi}_L \dot{\hat{\chi}}_L
\end{aligned}
\end{equation}
From \textit{Assumption II.1}, we define unknown constant $d^*_i \geq 0$ such that $|d_i| \leq d^*_i$. Thus, by inserting \eqref{3aadasdasd2},
\begin{equation}
\begin{aligned}
\small
\label{37}
\dot{\bar{V}} \leq & - \frac{1}{2} \epsilon_R e^2_R - {\gamma_R e^2_R [\log (\frac{{O}}{O-{E}})]^2 \hat{\chi}_R} + |e_R| \hspace{0.15cm} d^*_R\\
&-\frac{1}{2}\delta_R \hat{\chi}^2_R + {\gamma_R e^2_R [\log (\frac{{O}}{O-{E}})]^2 \hat{\chi}_R}\\
&- \frac{1}{2} \epsilon_L e^2_L - {\gamma_L e^2_L [\log (\frac{{O}}{O-{E}})]^2 \hat{\chi}_L} + |e_L| \hspace{0.15cm} d^*_L\\
&-\frac{1}{2}\delta_L \hat{\chi}^2_L + {\gamma_L e^2_L [\log (\frac{{O}}{O-{E}})]^2 \hat{\chi}_L}
\end{aligned}
\end{equation}
By applying the Cauchy-Schwarz inequality, we have
\begin{equation}
\begin{aligned}
\small
\label{37}
\dot{\bar{V}} \leq & - \frac{1}{2} \epsilon_R e^2_R + \frac{1}{2} \kappa_R e^2_R +  \frac{1}{2\kappa_R} {d^*_R}{^2} -\frac{1}{2}\delta_R \hat{\chi}^2_R\\
& - \frac{1}{2} \epsilon_L e^2_L + \frac{1}{2} \kappa_L e^2_L +  \frac{1}{2\kappa_L} {d^*_L}{^2} -\frac{1}{2}\delta_L \hat{\chi}^2_L
\end{aligned}
\end{equation}
where $\epsilon_i$ and $\kappa_i$ are positive constant such that $\epsilon_i > \kappa_i $. Thus,

\begin{equation}
\begin{aligned}
\small
\label{37}
\dot{\bar{V}} \leq & - \frac{1}{2} (\epsilon_R - \kappa_R) e^2_R  +  \frac{1}{2\kappa_R} {d^*_R}{^2} -\frac{1}{2}\delta_R \hat{\chi}^2_R\\
& \frac{1}{2} (\epsilon_L - \kappa_L) e^2_L  +  \frac{1}{2\kappa_L} {d^*_L}{^2} -\frac{1}{2}\delta_L \hat{\chi}^2_L
\end{aligned}
\end{equation}
Finally, we have $\dot{\bar{V}} \leq - \mu {\bar{V}} + \ell $ where $\mu=\min[A^{-1}_R (\epsilon_R - \kappa_R), A^{-1}_L (\epsilon_L - \kappa_L), \delta_R, \delta_L]$ and $\ell = \frac{1}{2\kappa_R} {d^*_R}{^2} + \frac{1}{2\kappa_L} {d^*_L}{^2}$. It guarantees that based on Definition 1 of \cite{shahasaasca2024asdacvrobustness}, the proposed RSDNN-based control guarantees uniform exponential stability of the black-box actuation mechanisms, while the safety module ensures the whole operation safety. 

\section{Experimental Validity}
The studied LSMR is a 6,000-kg skid-steering robot with four off-road wheels. It was equipped with two Basler acA1920-50gc global-shutter color cameras (30 fps), a Trimble BD992-INS in dual-antenna mode (20 Hz), and Danfoss EMD speed sensors on the in-wheel hydraulic motors (Fig. \ref{dnn_contcomdsfsdf}). Stereo calibration yields a 0.32 m baseline, and both cameras are pitched about $5^{\circ}$ downward to improve near-field coverage and reduce long-range false detections. Images are stored uncompressed, which ORB-SLAM3 expects, and global shutters reduce motion distortion on moving platforms. The BD992-INS provides the ORB-SLAM3 reference, with INS-RTK accuracy of 0.05 m horizontal, 0.03 m vertical, and orientation accuracy of $0.10^{\circ}$ roll/pitch and $0.09^{\circ}$ heading with a 2 m antenna baseline. Camera intrinsics were obtained via checkerboard calibration using Kalibr's Multi-Camera tool \cite{rehder2016extending}.
Hardware centers on an embedded Beckhoff PC that acquires sensors over EtherCAT, time-stamps them, and distributes UDP datagrams to the Linux perception stack. Camera streams use ROS drivers over UDP.

\begingroup
\setlength{\tabcolsep}{3pt}
\renewcommand{\arraystretch}{0.9}
\begin{table}[h]
\centering
\caption{Parameters for ORB-SLAM3 with setup depth 40.}
\footnotesize
\begin{tabular}{@{}ccccccc@{}}
\toprule
\shortstack{nFeat\\x1000} & \shortstack{FAST\\ini} & \shortstack{APE KF\\(mm)}
& \shortstack{APE Odom.\\(mm)} & \shortstack{RPE Odom.\\(mm)} & \shortstack{CPU\\(\%)} & \shortstack{Mem\\(\%)} \\
\midrule
1 & 12 & 590.6 & 513.8 & 110.8 & 14.21 & 3.41 \\
2 & 12 & 549.6 & \textcolor{darkergreen}{490.7} & 112.7 & 13.39 & 3.11 \\
3 & 12 & 599.1 & 557.2 & 130.7 & 12.73 & 3.32 \\
1 & 20 & 582.8 & 510.2 & 109.8 & 14.26 & 3.25 \\
2 & 20 & 562.9 & 508.3 & \textcolor{darkergreen}{109.6} & \textcolor{darkergreen}{12.58} & 2.88 \\
3 & 20 & 546.5 & 504.6 & 110.1 & 13.69 & 3.16 \\
\bottomrule
\end{tabular}
\label{asdbassadadkjb}
\end{table}
\endgroup


\begingroup
\setlength{\tabcolsep}{3pt}
\renewcommand{\arraystretch}{0.9}
\begin{table}[h]
\centering
\caption{Parameters for ORB-SLAM3 with setup depth 50.}
\footnotesize
\begin{tabular}{@{}ccccccc@{}}
\toprule
\shortstack{nFeat\\x1000} & \shortstack{FAST\\ini} & \shortstack{APE KF\\(mm)}
& \shortstack{APE Odom.\\(mm)} & \shortstack{RPE Odom.\\(mm)} & \shortstack{CPU\\(\%)} & \shortstack{Mem\\(\%)} \\
\midrule
1 & 12 & 604.5 & 509.1 & 109.9 & 14.02 & 3.46 \\
2 & 12 & 552.6 & 510.9 & 111.4 & 12.39 & \textcolor{darkergreen}{2.26} \\
3 & 12 & 643.8 & 570.0 & 130.1 & 12.66 & 3.20 \\
1 & 20 & 598.4 & 516.6 & 109.6 & 14.02 & 3.52 \\
2 & 20 & \textcolor{darkergreen}{542.8} & 502.8 & 109.7 & 13.19 & 3.14 \\
3 & 20 & 556.0 & 498.2 & 109.7 & 13.09 & 3.13 \\
\bottomrule
\end{tabular}
\label{asdbasdkjb}
\end{table}
\endgroup
Stereo ORB-SLAM3 was tuned offline on rough terrain, then validated online. Trajectories were compared to INS-RTK ground truth with evo on Ubuntu 20.04 and ROS Noetic using an i9-14900K and 32 GB RAM. Twelve configurations were tested while tracking compute load, mapping quality, pose accuracy, and loop-closure reliability. The most influential parameters were per-frame feature count, FAST thresholds, and the stereo depth cutoff, about 12.8 m at 40 and 16.0 m at 50. Accuracy was reported as absolute pose error (APE) and relative pose error (RPE).
Best results occurred at about 2000 features, with APE slightly worse at 1000 or 3000. Set 2 (depth 40, FAST 12 px) raised candidate counts and still gave reliable estimates despite mildly lower ORB quality. The deeper 50 with FAST 20 was also strong, yielding higher quality points and longer range. Loop closure succeeded in all cases with no visible Z drift before correction. CPU peaked near 1000 features, eased slightly as features increased, averaging about 13 percent. Memory stayed near 3 percent across configurations. As shown in Fig. \ref{i9sasddadado_synthesnbfgfhfhnfh}, ORB-SLAM3 publishes the robot pose $\boldsymbol{x}_{\mathrm{msr}}(t)$ at 20 Hz to the high-level NMPC, and the in-wheel speed sensors provide $\dot{\boldsymbol{\theta}}_{\text {msr }}(t)$ at 1 kHz.
The NMPC computes optimal wheel commands $\dot{\boldsymbol{\theta}}(t)$ every 1 ms and sends them to the low-level controller, which regulates two-side PMSMs' RPMs to rotate the demanding in-wheel hydraulic motor speed $\dot{\boldsymbol{\theta}}_{\text {msr }}(t)$ in order to achieve the updated wheel velocity (see Fig. \ref{dnn_contcomdsfsdf}). All computation runs onboard within the Beckhoff system. In other words, the high-level controller drives $\boldsymbol{x}_{\mathrm{msr}}(t)$ toward the reference $\boldsymbol{x}^r(t)$ by generating updated wheel-speed commands $\dot{\boldsymbol{\theta}}(t)$. The low-level controller handles actuation, regulating the PMSM motors $u_R (t)$ and $u_L (t)$ so that $\dot{\boldsymbol{\theta}}_{\text {msr }}(t) \to \dot{\boldsymbol{\theta}}(t)$. Implemented in C++ with a nonlinear optimizer, the NMPC used symbolic expressions, BFGS \cite{nocedal2006numerical}, and warm starts to solve within sensor sampling periods. A refined warm start from an initial high-accuracy phase seeded the online solves. Real-time was ensured by a small fixed iteration budget, with frequent NMPC updates refining solutions across steps \cite{grandia2023perceptive}. UDP and smart-pointer buffering minimized queueing and latency.
All experiments ran on an industrial PC (Nuvo-9160GC) with an Intel Core i9, NVIDIA RTX 3050, 32 GB RAM, and a 1 TB SSD, using Linux with a C++ implementation. The unit was mounted on the mobile platform and networked via Ethernet. The NLP was developed in MATLAB with CasADi for C++ code generation, supplying the problem and first-order derivatives to IPOPT. BOOST handled UDP communication. Low-level control ran on a Beckhoff CX2043 at a 1 ms sampling rate. 
With warm starts and high-rate sensors, each NMPC step executes a single SQP iteration while still improving the NLP \cite{grandia2023perceptive}. We set $R=\operatorname{diag}(0.2,0.2)$ and $Q_x=Q_{x N}=\operatorname{diag}(20,20,12)$. The high-level NMPC bounds are
$\dot{\boldsymbol{\theta}}_{\min }=[0,0]$, $\dot{\boldsymbol{\theta}}_{\max }=[0.8,0.8]$, $\ddot{\boldsymbol{\theta}}_{\min }=[-0.2,-0.2]$, and $\quad \ddot{\boldsymbol{\theta}}_{\max }=[0.2,0.2]$.
Note that we did not consider any robot pose constraint $(\boldsymbol{x}_{\min }=[-\mathrm{inf},-\mathrm{inf},-\mathrm{inf}]$ and $\boldsymbol{x}_{\max }=[\mathrm{inf}, \mathrm{inf}, \mathrm{inf}]) $ for computational reduction to solve NLP; however, $\boldsymbol{x}_{\mathrm{msr}}$ is further limited by the safety-defined performance module; see Eq. \eqref{3asdsad2}.

For the SDNN module within the low-level control framework, we collected $9,900,000$ synchronized samples by perturbing the PSMMs' RPM signals $u_{\text{SDNN}_i}$ and logging the resulting track velocities $v_i$ with Danfoss EMD sensors across varied conditions. Data streams were aligned, de-spiked, and split into training, validation, and test sets.
From the raw vectors, indices were partitioned 70/15/15 using MATLAB's \textit{dividerand}. The training subset was scaled to $[-1,1]$ with \textit{mapminmax}, and the same parameters were applied to validation and test.
We trained MATLAB SDNNs with five hidden layers $[35,20,12,10,8]$, tan-sigmoid hidden activations, and a linear output, with built-in preprocessing disabled. Optimization used scaled conjugate gradient (\textit{trainscg}). Targets were MSE $1 \times 10^{-3}$, minimum gradient $1 \times 10^{-4}$, and a 200-epoch cap, with early stopping based on validation MSE.

\begin{figure}[h!]
\hspace*{-0.0cm} 
\centering
\scalebox{0.8}{\includegraphics[trim={0cm 0.0cm 0.0cm 0cm},clip,width=\columnwidth]{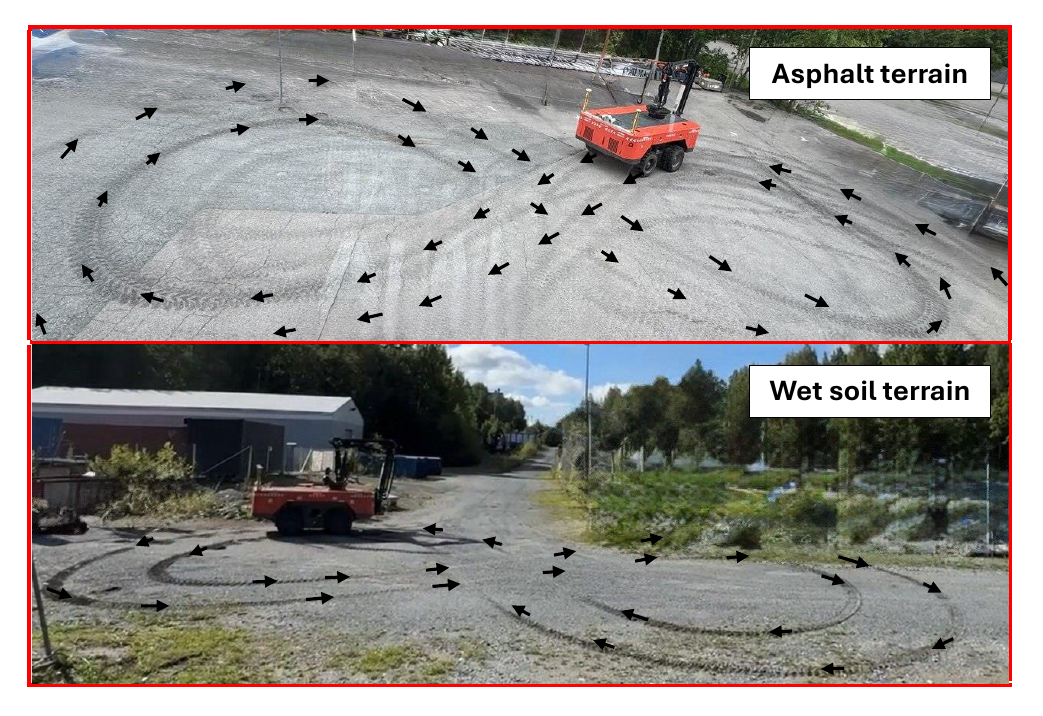}}
\caption{Real-world validation on asphalt and soft soil.}
\label{Realworldvalidationssad}
\end{figure}

\begin{figure}[h!]
\hspace*{-0.0cm} 
\centering
\scalebox{0.7}{\includegraphics[trim={0cm 0.0cm 0.0cm 0cm},clip,width=\columnwidth]{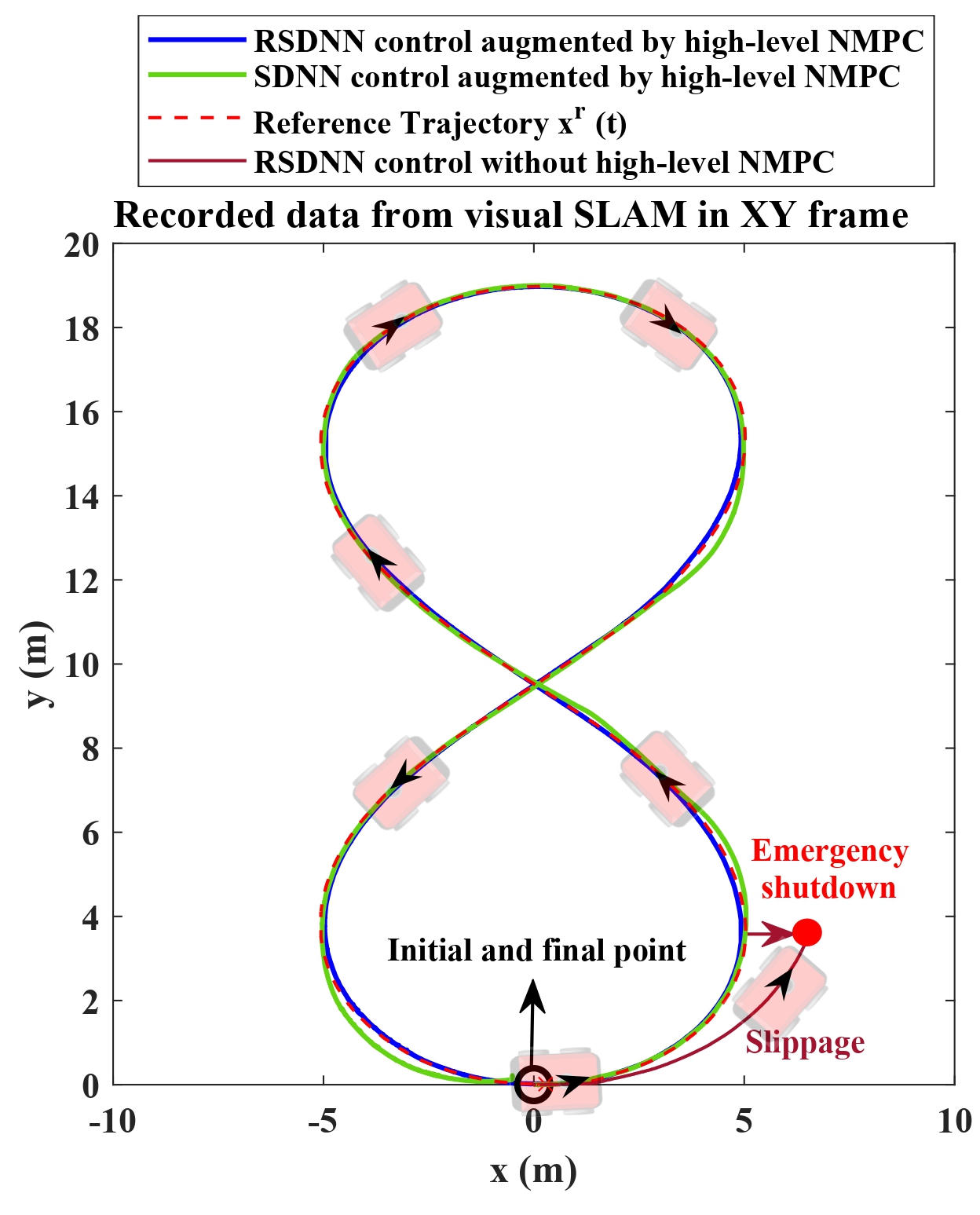}}
\caption{SLAM data for three scenarios on the soft soil terrain.}
\label{ecordedslamm}
\end{figure}

To rigorously evaluate, as shown in Figs. \ref{Realworldvalidationssad}, and \ref{ecordedslamm}, we evaluated multiple configurations on two ground types: asphalt and soft soil. On asphalt, where the LSMR exhibited only minor slip, the results were nearly identical. On soft soil, which is a common surface for off-road LSMRs, the results diverged markedly, yielding substantial performance differences. To evaluate the whole framework, three configurations were implemented for the robot to track a reference pose trajectory, a lemniscate 19 m long and 10 m wide, as follows.

\begin{figure}[h!]
\hspace*{-0.0cm} 
\centering
\scalebox{0.9}{\includegraphics[trim={0cm 0.0cm 0.0cm 0cm},clip,width=\columnwidth]{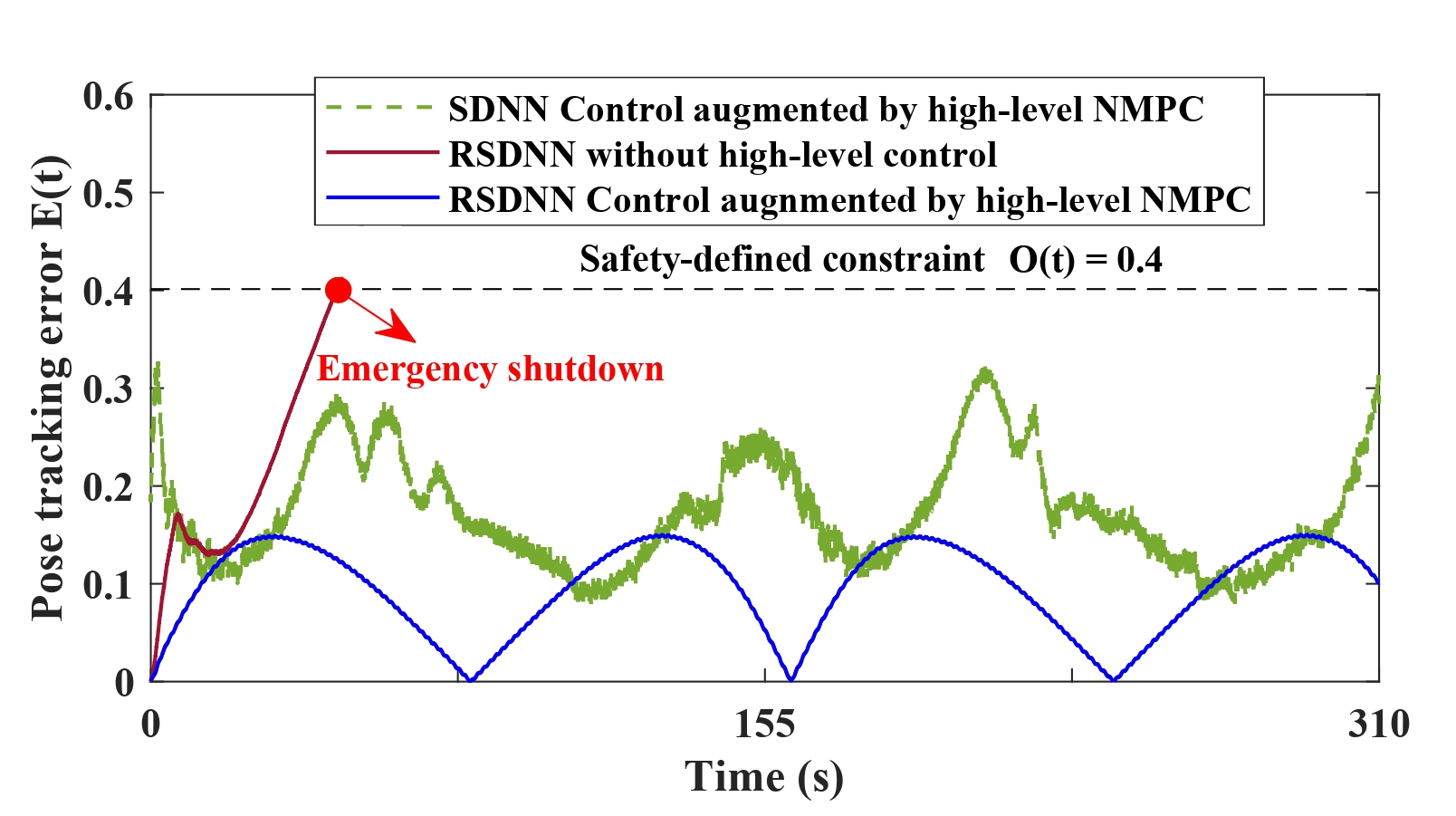}}
\caption{Results of the logarithmic module (E(t) in meters).}
\label{slamerrorlog}
\end{figure}

\textit{(1) Absence of high-level control}: We executed the full proposed framework with the high-level NMPC disabled. The wheel reference velocities, analytically derived from the predefined lemniscate trajectory ${\bm{x}}^r(t)$, were sent directly to the low-level controller (RSDNN policy) in the actuation layer. This scenario was designed to demonstrate the role of the high-level controller under slippage.
On asphalt, the task was completed successfully even without high-level NMPC, due to minimal slippage and strong wheel-ground contact. On soft soil, although the low-level loop accurately tracked the actuator-level velocity commands, terrain-induced slippage of varying intensity caused progressive pose drift relative to the robot-level reference. Once the drift exceeded the predefined safety bounds $E(t) < O = 0.4$, the proposed logarithmic safety module triggered an emergency shutdown (see the dark red trajectories in Figs. \ref{ecordedslamm} and \ref{slamerrorlog}). The results of this scenario confirm the necessity of the proposed high-level controller for LSMRs.

\textit{(2) SDNN low-level control in the actuation mechanism}: We deployed the full framework, including visual SLAM for pose estimation, the high-level NMPC that updated the reference commands in real time under slippage, and the trained SDNN low-level control policy in the actuation mechanism, but without the adaptive law to handle potential out-of-distribution behavior (see the green trajectories in Figs. \ref{ecordedslamm} and \ref{slamerrorlog}). The results were largely successful within the safety-defined constraints, with minor inaccuracies at points where the robot had to change orientation sharply, especially near locations (5, 6) and (13, 5). These errors occurred because rapid orientation changes pushed the actuation mechanism into out-of-distribution conditions for the trained SDNN model, since the adaptive controller was disabled.

\textit{(3) Full proposed framework}: We deployed the complete framework, comprising visual SLAM for pose estimation, a high-level NMPC that updates the reference commands in real time under slippage, and the RSDNN actuation policy, which combines the trained SDNN controller with an adaptive law to handle potential out-of-distribution behavior. The framework performed robustly and remained within the safety-defined constraints (see the blue trajectories in Figs. \ref{ecordedslamm} and \ref{slamerrorlog}).

\begin{figure}[h!]
\hspace*{-0.0cm} 
\centering
\scalebox{1}{\includegraphics[trim={0cm 0.0cm 0.0cm 0cm},clip,width=\columnwidth]{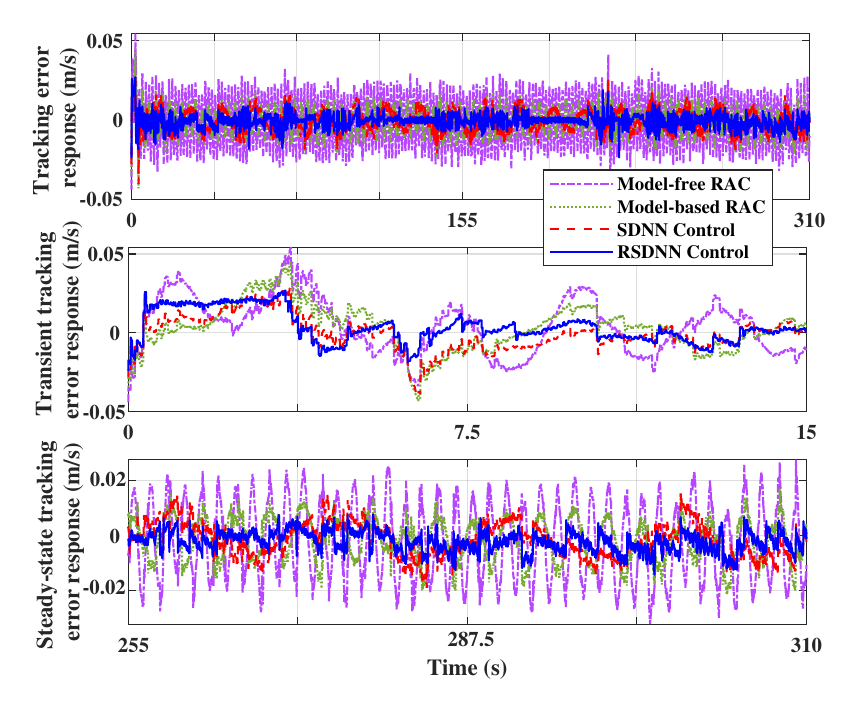}}
\caption{Actuation-level tracking across three time spans.}
\label{low-leveltrack}
\end{figure}

In addition to the comparative study of the full framework, we also implemented two state-of-the-art robust controllers at the actuation level for comparison with the proposed policies: a model-based robust adaptive controller (RAC) augmented with neural networks \cite{chen2021model}, and a model-free RAC \cite{shahna2025model}. This evaluation ignored the robotic task and focused solely on regulating PMSM RPM inputs to generate sufficient wheel motion to track the wheel-motion references, despite pronounced nonlinearities and the complexity of the multi-stage actuation mechanism (see Fig. \ref{dnn_contcomdsfsdf}). As shown in Fig. \ref{low-leveltrack} and Table \ref{asdbassadasadadadkjb}, all four controllers tracked the reference wheel motions with varying accuracy.

The RSDNN policy delivered the strongest overall performance: the shortest peak time, the smallest maximum overshoot, a short settling time, and the lowest steady-state error. SDNN was a close second, with the fastest settling time. The model-based RAC ranked next. The model-free RAC performed worst, exhibiting higher-frequency oscillations and larger tracking errors, likely because its model-free nature requires stronger adaptation to compensate for unknown dynamics. Although the model-based RAC occasionally matched the proposed policies in peak and settling time, its accuracy errors would compound with those from other modules in challenging robotic tasks, leading to tangible performance losses. Therefore, the proposed higher-accuracy actuation-level control policy is valuable for the overall framework.

\begingroup
\setlength{\tabcolsep}{3pt}
\renewcommand{\arraystretch}{0.9}
\begin{table}[h]
\centering
\caption{Summarized Comparison of different low-level control methods in the actuation level.}
\footnotesize
\begin{tabular}{@{}ccccc@{}}
\toprule
\shortstack{Control\\method} & \shortstack{Peak\\time (s)} & \shortstack{Maximum \\overshoot (m/s)}
& \shortstack{Settling\\time (s)} & \shortstack{Steady-state\\error (m/s)} \\
\midrule
Model-free RAC \cite{shahna2025model} & 3.050 & 0.060 & 8.150 & 0.021 \\
Model-based RAC \cite{chen2021model}  & 3.200 & 0.040 & 4.650 & 0.019 \\
SDNN control policy & 3.100 & 0.029 & \textcolor{darkergreen}{4.500} & 0.017 \\
RSDNN control policy & \textcolor{darkergreen}{3.000} & \textcolor{darkergreen}{0.025} & 4.550 & \textcolor{darkergreen}{0.008} \\
\bottomrule
\end{tabular}
\label{asdbassadasadadadkjb}
\end{table}
\endgroup

\section{Conclusion}
This paper has presented a comprehensive control framework for a 6,000 kg LSMR that ensures stability and safety-defined performance and enables robust operation on slippery off-road terrain. First, ORB-SLAM3 in a stereo camera configuration has estimated the robot’s pose, which the high-level NMPC has used to update wheel-motion commands in real time to compensate for off-road wheel slip. Then, the RSDNN control policy at the hybrid actuation level has adjusted the in-wheel PMSMs and hydraulic motors so that the wheels tracked the updated commands. A logarithmic safety module has ensured closed-loop safety and the specified performance. Experimental results have validated the framework across multiple scenarios and comparative configurations.
To our knowledge, this is the first study in the literature of a multi-thousand-kilogram LSMR that achieved real-world autonomy by integrating high-performance deep learning while guaranteeing system stability and safety on off-road terrain.

\bibliographystyle{IEEEtran}

\bibliography{MEHDI}

\vfill

\end{document}